\documentclass[conference]{IEEEtran}
\IEEEoverridecommandlockouts

\usepackage{amsmath,amssymb,amsfonts}
\usepackage{graphicx}
\usepackage{color}
\usepackage{multirow}
\usepackage{hyperref}

\newcommand\etal{{\it et~al.}}
\newcommand\ie{{\it i.e.}}

\newcommand{\pink}[1]{\textcolor{magenta}{#1}}
\def\BibTeX{{\rm B\kern-.05em{\sc i\kern-.025em b}\kern-.08em
    T\kern-.1667em\lower.7ex\hbox{E}\kern-.125emX}}
\begin{document}

\title{Minutes to Seconds: Speeded-up DDPM-based Image Inpainting with Coarse-to-Fine Sampling\\
}


\author{\IEEEauthorblockN{Lintao Zhang$^{1,2}$, Xiangcheng Du$^{1,3}$, LeoWu TomyEnrique$^{1}$, Yiqun Wang$^{1}$, Yingbin Zheng$^{3*}$, Cheng Jin$^{1,2*}$\thanks{$^{*}$Corresponding author.}}
\IEEEauthorblockA{$^1$\textit{School of Computer Science, Fudan University, Shanghai, China}\\
$^2$\textit{Innovation Center of Calligraphy and Painting Creation Technology, MCT, China}\\
$^3$\textit{Videt Technology, Shanghai, China}\\
    \{ltzhang21, xcdu22, yiqunwang23\}@m.fudan.edu.cn,\\ enriqueliu1999@gmail.com, zyb@videt.cn, jc@fudan.edu.cn
    }    
}

\maketitle

\begin{abstract}
For image inpainting, the existing Denoising Diffusion Probabilistic Model (DDPM) based method \ie~RePaint can produce high-quality images for any inpainting form. 
It utilizes a pre-trained DDPM as a prior and generates inpainting results by conditioning on the reverse diffusion process, namely denoising process.
However, this process is significantly time-consuming.
In this paper, we propose an efficient DDPM-based image inpainting method which includes three speed-up strategies.
First, we utilize a pre-trained Light-Weight Diffusion Model (LWDM) to reduce the number of parameters.
Second, we introduce a skip-step sampling scheme of Denoising Diffusion Implicit Models (DDIM) for the denoising process.
Finally, we propose Coarse-to-Fine Sampling (CFS), which speeds up inference by reducing image resolution in the coarse stage and decreasing denoising timesteps in the refinement stage.
We conduct extensive experiments on both faces and general-purpose image inpainting tasks, and our method achieves competitive performance with approximately 60 times speedup.
The source code and trained models are available at \pink{\href{https://github.com/linghuyuhangyuan/M2S}{https://github.com/linghuyuhangyuan/M2S}}.
\end{abstract}


\section{Introduction}
\label{sec:intro}
Image inpainting aims to fill missing or damaged regions in an image with reasonable visual content~\cite{bertalmio2000image}. 
The technology can be used in many applications, such as restoring damaged photos~\cite{wan2020bringing}, editing images~\cite{jo2019sc}, removing objects~\cite{du2023progressive}, and eliminating unwanted objects from images~\cite{barnes2009patchmatch}.

Recently, GAN-based~\cite{zhao2021large,wang2022dual} or Autoregressive Modeling based~\cite{peng2021generating,yu2021diverse} approaches achieve superior performances.
However, these methods struggle with novel mask types due to the specific mask distributions in the training stage.
Diffusion-based methods change the inference process instead of training one specific conditional DDPM for image inpainting. The strategy allows the model to produce high-quality and diverse output images for any inpainting form.
For example, RePaint~\cite{lugmayr2022repaint} employs DDPM~\cite{ho2020denoising} as a generative prior and is applicable to free-form inpainting with any mask type. It performs better on masks of different distributions compared with mask-specific training approaches. However, the inference process using DDPM demands a great amount of time.

\begin{figure}[t]
  \centering
  \includegraphics[width=\linewidth]{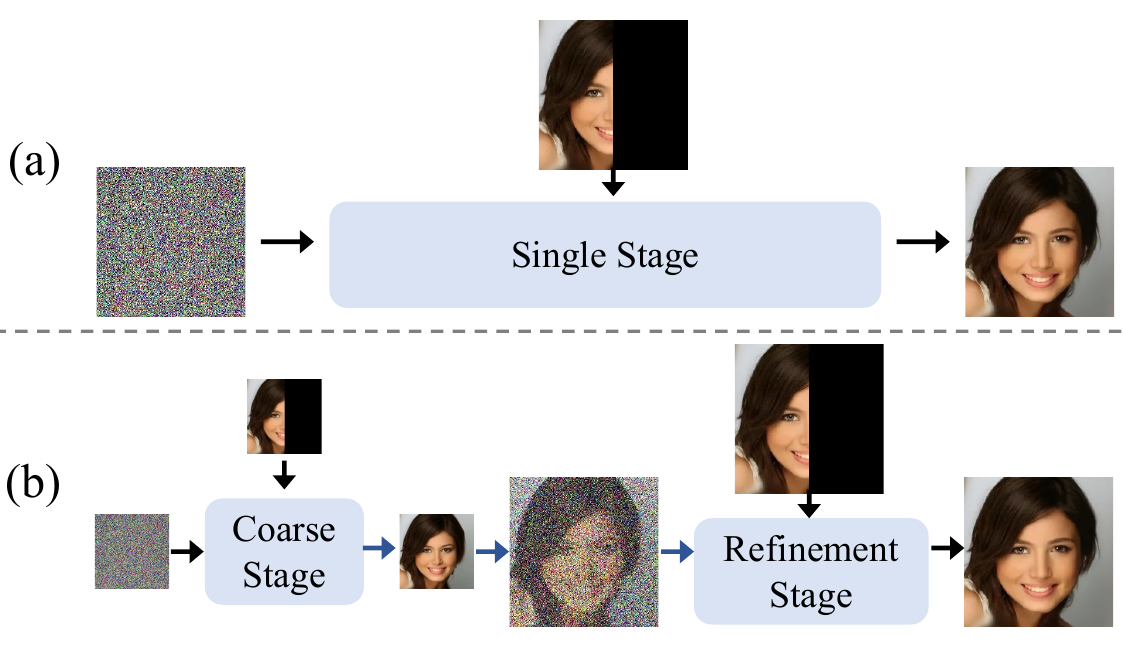}
  \caption{(a) Single-stage DDPM-based inpainting method RePaint~\cite{lugmayr2022repaint}. (b) Our proposed efficient DDPM-based method using Coarse-to-Fine Sampling (CFS).}
  \label{fig:fig1}
\end{figure}

In this paper, we propose three strategies to speed up the single-stage method~\cite{lugmayr2022repaint}. 
As shown in Fig.~\ref{fig:fig1}(a), the single-stage method directly samples the inpainting result through the denoising process with the masked input as condition.
In our method, we first produce the coarse result from a random noise by the conditioned coarse stage. Then the result is upsampled and noise-added to serve as the input of the refinement stage and generate the refined result. In each stage, we introduce Conditioned Denoising Module (CDM) to denoise, and Conditioned Resampling Module (CRM) make the inpainting result more harmonious.

Based on the framework, our three strategies are explained as follows.
Due to the large parameters in the original pre-trained diffusion model~\cite{dhariwal2021diffusion}, 
we substitute it with a Light-Weight Diffusion Model (LWDM) by reducing model parameters combined with the modified objective function. Following~\cite{choi2022perception}, we adjust the weighting scheme of the training loss, which aims to prioritize learning from more important noise levels and can compensate for parameter reduction.
Additionally, skip-step DDIM sampling~\cite{song2020denoising} replaces DDPM sampling in CDM. 
Moreover, we propose Coarse-to-Fine Sampling (CFS) to split the denoising process into two time-short stages. In the coarse stage where the reverse diffusion process is long due to large denoising timesteps, we lower the image resolution and reduce the number of CRMs. With the given image prior from the coarse stage, the refinement stage acquires fewer denoising steps and we can apply more CRMs to enhance image details.


The main contributions of our work are as follows:
\begin{itemize}
    \item We introduce a loss-redesigned Light-Weight Diffusion Model which demonstrates a remarkable acceleration with a minor impact on inpainting performance.
    \item We propose a dual-stage method using Coarse-to-Fine Sampling with DDIM sampling, demonstrating fast speed and excellent performance.
    \item Extensive experiments show that our method can produce competitive performance compared to the DDPM-based method with about 60$\times$ speedup.
\end{itemize}

\section{Related Work}
\label{sec:related}
Traditional inpainting methods can be mainly divided into two categories, \ie~diffusion-based and patch-based methods. Diffusion-based methods~\cite{ballester2001filling, bertalmio2000image} render masked regions referring to the appearance information of the neighboring undamaged ones. Patch-based methods~\cite{barnes2009patchmatch, xu2010image} reconstruct masked regions by searching and pasting the most similar patches from undamaged regions of images.
Although these methods achieve good performance, they have high computational cost in calculating patch similarities, and are difficult to reconstruct patches with rich semantics.

Recently, the deep generative methods have achieved great successes on the image inpainting task. To produce much sharper results, Pathak~\etal~\cite{pathak2016context} introduce the adversarial loss in image inpainting and utilize context encoder to learn the semantics of visual structures.
Nazeri~\etal~\cite{nazeri2019edgeconnect} propose EdgeConnect that fills in the missing regions using hallucinated edges as prior.
In order to address modeling long-range interactions in the inpainting problem, Li~\etal~\cite{li2022mat} purpose MAT for large hole inpainting, which unifies the merits of transformers and convolutions.
Most existing methods train for a certain distribution of masks, Lugmayr~\etal~\cite{lugmayr2022repaint} propose a novel conditioning approach named RePaint, which complies with the assumptions of DDPM and increases the degree of freedom of masks for the free-form inpainting.

\begin{figure*}[t]
    \centering
    \includegraphics[width=\linewidth]{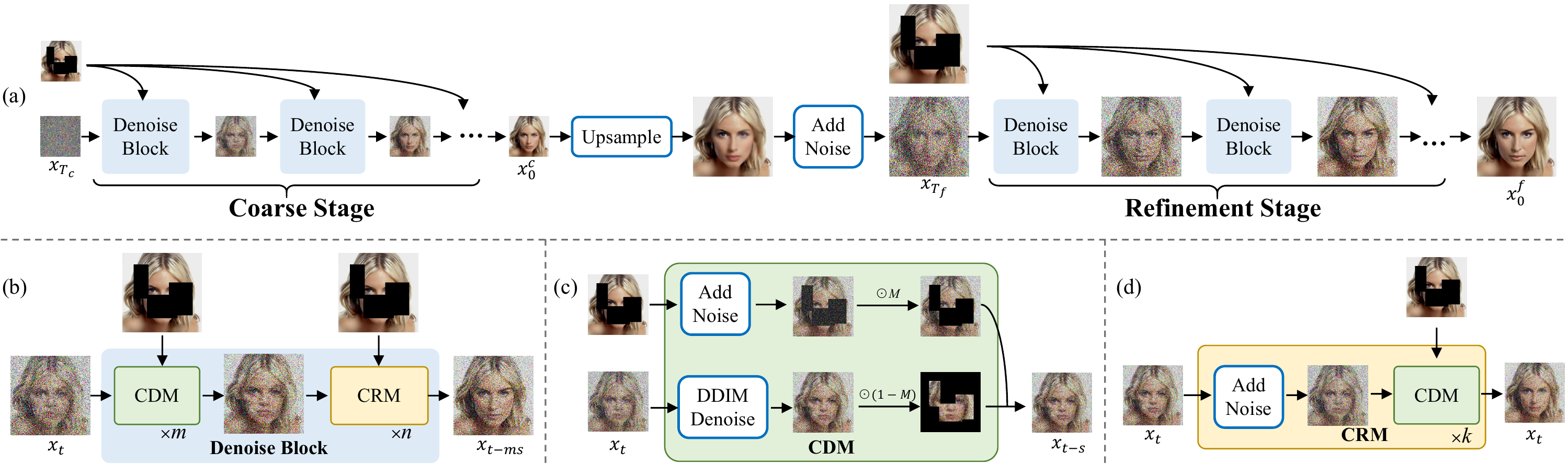}
    \caption{Framework of our method. (a) We sample the final result $x_0^f$ from a random Gaussian noise $x_{T_c}$ through two-stage reverse diffusion process with condition guided. 
    (b) Denoise Block denoises $x_t$ to $x_{t-ms}$, which includes $m$ CDMs and $n$ CRMs.
    (c) Conditioned Denoising Module (CDM) denoises $x_t$ to $x_{t-s}$ with condition added.
    (d) Conditioned Resampling Module (CRM) converts $x_t$ to more harmonious $x_t$ with the fusion of conditioning information and generated content.}
    \label{fig:pipline}
  \end{figure*}

\section{Background}
\label{sec:background}

\textbf{Denoising Diffusion Probabilistic Model} (DDPM)~\cite{ho2020denoising} is another generative model using Markov chain to transform latent variables in simple distributions to complex data distributions.
A diffusion model is composed of two processes: \textbf{diffusion process} and \textbf{reverse process}.

The \textbf{diffusion process} is a Markov chain that adds gaussian noise to data:
\begin{equation}
    q(x_t|x_{t-1})= \mathcal{N}(x_t;\sqrt{1-\beta _t}x_{t-1},\beta _tI)
\end{equation}
where $ \beta _t \in (0, 1)$ for all $t = 1, . . . , T$.
Using the reparameterization trick, the diffusion process allows sampling $x_t$ at any given timestep $t$ in closed form: 
\begin{equation} 
    q(x_t|x_0) = \mathcal{N}(x_t; \sqrt{\bar\alpha_t}x_0, (1-\bar\alpha_t)\textbf{I}),
    \label{q_marginal_arbitrary_t} 
\end{equation} 

The \textbf{reverse process} is a Markov chain that converts noise back into data distribution:
\begin{equation}
p_{\theta }(x_{t-1}|x_t)= \mathcal{N}(x_{t-1};\mu _\theta (x_t,t),\Sigma _{\theta}(x_t,t))
\label{p_process}
\end{equation}
where $\theta$ denotes model parameters, and the mean or variance is parameterized by this model.

In inference, we can generate data directly from Gaussian noise. We sample an $x_T \sim \mathcal{N} (x_T;\textbf{0}, \textbf{I})$, and then sample $x_{t-1} \sim p_\theta (x_ {t-1}|x_t)$ according to Eq.~\eqref{p_process}. Finally, we obtain predicted origin image $x_0$ through a continuous sample.

DDPM is proved to be effective for generating high-quality and diverse results~\cite{dhariwal2021diffusion,ho2022classifier}.
Lugmayr~\etal~\cite{lugmayr2022repaint} first introduced DDPM-based method for image inpainting.
In this paper, we also employ a pre-trained unconditional DDPM (LWDM) and introduce the masked input image to the reverse diffusion process to condition generating.
In each denoising step, by Eq.~\eqref{p_process}, we can obtain the denoised result $x_{t-1}^\text{unknown} \sim \mathcal{N}(\mu _\theta (x_t,t),\Sigma _{\theta}(x_t,t))$, which includes the generated infomation $M \odot x_{t-1}^\text{unknown}$.
Meanwhile, we use Eq.~\eqref{q_marginal_arbitrary_t} to sample the aligned known infomation $x_{t-1}^\text{known}$ from the masked input image.
Finally, we use the concatenated result $x_{t-1}$ as the input of the next denoising step:
\begin{equation}
    x_{t-1} = M \odot x_{t-1}^\text{known} + (1-M) \odot x_{t-1}^\text{unknown}
    \label{eq:concat}
\end{equation}
where $M$ represents the binary mask, and $\odot$ denotes element-wise product.


\section{Methodology}
\label{sec:method}
\subsection{Overview}

Given the masked input image as condition, we sample the final inpainted result from a random Gaussian noise through two-stage reverse diffusion process. The framework is illustrated in Fig.~\ref{fig:pipline}(a). In the coarse stage, we need $T_c$ denoising timesteps to sample coarse result $x_0^c$ from a random Gaussian noise $x_{T_c}$ by conditioned Denoise Blocks. The upsampled coarse result serves as the image prior for the refinement stage, where upsampling employs a simple bilinear interpolation method.  In the refinement stage, the input $x_{T_f}$ is sampled from the upsampled result at the specified timestep $T_f$ using Eq. \ref{q_marginal_arbitrary_t}. The refined final result $x_0^f$ is then sampled from it.   

For these two stages, we employ structure-consistent Denoise Blocks but with distinct parameter settings. In the Denoise Block, $m$ Conditioned Denoising Modules (CDM) are first applied to denoise $x_t$ to $x_{t-ms}$ while adding conditional information. Then $n$ Conditioned Resampling Modules (CRM) are employed to enhance the fusion of conditional information with the generated content. These two modules are explained in Section \ref{sec:cdm} and Section \ref{sec:crm}.

Our method is more efficient than RePaint \cite{lugmayr2022repaint} due to three speed-up strategies: LWDM, DDIM and CFS. 
Firstly, we replace the pre-trained DDPM model with a lightweight one to complete noise prediction during each denoising step, which is explained in Section \ref{sec:lwdm}.
Secondly, in CDM, DDIM Denoise is introduced, which supports skip-step sampling. 
Finally, the specially designed CFS demonstrates higher efficiency and superior results. In the coarse stage, we choose lower resolution and less CRM iterations because of the larger denoising timesteps $T_c$. In the refinement stage, $T_f$ is small with the image prior, so we can apply more CRMs to obtain more harmonious detail-refined results. 
Our method offers significantly improved speed with competitive inpainting performance.

\subsection{Light-Weight Diffusion Model}
\label{sec:lwdm}
Intuitively, reducing parameters of the diffusion model can accelerate inference process.
However, directly reducing parameters may compromise the performance of the model. Inspired by~\cite{choi2022perception}, we modify the loss function in training process of original DDPM to compensate for parameter reduction.

The original objective in~\cite{ho2020denoising} can be described as the following form:
\begin{equation}
    \mathcal{L}=\sum_t\lambda _t \mathcal{L}_t
\end{equation}
where weighting scheme $\lambda _t=(1-\beta _t)(1-\alpha _t)/\beta _t$ and $L_t$ is denosing score matching loss~\cite{vincent2011connection}.

The modified loss function introduces perception prioritized weighting, which emphasize training on the content stage to encourage the model to learn perceptually rich contexts. The $\lambda_t$ can be replaced as $\lambda'_t$:
\begin{equation}
    \label{eq:p2}
    \lambda'_t = \frac{\lambda_t}{(k+\text{SNR}(t))^{\gamma}},
\end{equation}
where SNR is signal-to-noise ratio, SNR of noisy image $x_t$ is $SNR(t)=\frac{\alpha_t}{1-\alpha_t}$. The modified loss function can be defined as $\mathcal{L}=\sum_t\lambda' _t L_t$.

\subsection{Conditioned Denoising Module}
\label{sec:cdm}
We introduce the conditional information in each denoising step. CDM is shown in Fig.~\ref{fig:pipline}(c). For the input image $x_t$ undergoing CDM, the resulting output $x_{t-s}$ is denoised and fused representation, where $s$ represents the stride of DDIM sampling. 

We employ the pre-trained LWDM and utilize the non-Markovian inference process of DDIM to obtain the denosied result $x_{t-s}^\text{known}$, which can be described as:
\begin{align}
    \begin{aligned}
        \mathbf{x}_{t-s}^\text{unknown} & = \sqrt{\bar{\alpha}_{t-s}} \left(\frac{\mathbf{x}_t - \sqrt{1 - \bar{\alpha}_t} \epsilon_\theta^{(t)}(\mathbf{x}_t)}{\sqrt{\bar{\alpha}_t}}\right) \\
        & \quad + \sqrt{1 - \bar{\alpha}_{t-s} - \sigma_t^2} \cdot \epsilon_\theta^{(t)}(\mathbf{x}_t) + \sigma_t \epsilon_t \label{eq:sample-eq-gen}
    \end{aligned}
\end{align}
where $\epsilon_t \sim \mathcal{N}(\mathbf{0}, \mathbf{I})$ is standard Gaussian noise independent of $\mathbf{x}_t$. DDIM sampling is a special case when $\sigma_t=0$.
Then, we align the known information with $x_{t-s}^\text{unknown}$ by diffusing the input image according to Eq.~\eqref{q_marginal_arbitrary_t}.
Finally, we concatenate the generated information $x_{t-s}^\text{unknown}$ with the conditional information $x_{t-s}^\text{known}$ as Eq.~\eqref{eq:concat}.

\subsection{Conditioned Resampling Module}
\label{sec:crm}
CDM aims to introduce the conditional information of the input image.
Moreover, we need CRM to harmonize the given information $M \odot x_t^\text{known}$ with the generated infomation $(1-M) \odot x_t^\text{unknown}$ by diffusing to $x_{t+ks}$ and denoising back to $x_t$.
The pipeline of CRM is shown in Fig.~\ref{fig:pipline}(d).
Specifically, CRM takes $x_t$ as input and produces resampled $x_t$ as output. 
In each iteration, $x_t$ is first noised for $ks$ steps to obtain $x_{t+ks} \sim \mathcal{N}(\sqrt{\bar{\alpha}_{t+ks}}\mathbf{x}_t, (1-\bar{\alpha}_{t+ks})\mathbf{I})$, and then denoised back to $x_t$ through $k$ CDMs. 
The original $x_t$ is then replaced with this denoised version.


\begin{table*}[t]
  \centering
  \caption{
  Comparison between different speed-up strategies. For DDIM Sampling, we fix the stride $s=5$. Without CFS means that we use the single-stage method. The best score is highlighted in \textbf{bold}.
  }
  \label{tab_ablation1}
  \small 
  \begin{tabular}{ccc|c|c|ccc|ccc} \hline
  \multicolumn{3}{c|}{\textbf{Strategies}} & \multirow{2}{*}{\begin{tabular}[c]{@{}c@{}}\textbf{Inference}\\ \textbf{Time}\end{tabular}} &
  \multirow{2}{*}{\begin{tabular}[c]{@{}c@{}}\textbf{Acceleration}\\ \textbf{Ratio}\end{tabular}} &
  \multicolumn{3}{c|}{\textbf{Wide}} & \multicolumn{3}{c}{\textbf{Narrow}}
  \\
  \cline{1-3} \cline{6-11}
  LWDM & DDIM & CFS & & & LPIPS$\downarrow$ & SSIM$\uparrow$ & $\boldsymbol{\ell_1(\%)}\downarrow$ & LPIPS$\downarrow$ & SSIM$\uparrow$ & $\boldsymbol{\ell_1(\%)}\downarrow$ \\ \hline
   & & & $938.2s$ & $-$ & 0.0706 & 0.8771 & 2.51 & 0.0353 & \textbf{0.9210} & 1.15 \\
   & & $\checkmark$ & $395.7s$ & $2.4 \times$ & \textbf{0.0691} & 0.8763 & 2.58 & \textbf{0.0343} & 0.9196 & \textbf{1.14} \\
  $\checkmark$ & & & $189.4s$ & $4.9 \times$ & 0.0719 & 0.8796 & 2.48 & 0.0364 & 0.9186 & 1.15 \\
   & $\checkmark$ & & $186.6s$ & $5.0 \times$ & 0.0791 & 0.8778 & 2.69 & 0.0397 & 0.9191 & 1.21 \\
  $\checkmark$ & & $\checkmark$ & $76.3s$ & $12.3 \times$ & 0.0693 & 0.8820 & \textbf{2.35} & 0.0374 & 0.9166 & 1.18 \\
  & $\checkmark$ & $\checkmark$ & $75.5s$ & $12.4 \times$ & 0.0709 & 0.8811 & 2.45 & 0.0371 & 0.9161 & 1.18 \\
  $\checkmark$ & $\checkmark$ & & $37.9s$ & $24.8 \times$ & 0.0787 & 0.8814 & 2.57 & 0.0417 & 0.9182 & 1.22 \\
  $\checkmark$ & $\checkmark$ & $\checkmark$ & $15.4s$ & $60.9 \times$ &0.0724 & \textbf{0.8834} & 2.37 & 0.0393 & 0.9174 & 1.20 \\ \hline
  \end{tabular}
  \end{table*}

\begin{table*}[t]
  \centering
  \caption{Comparison between different designs of CFS strategies. 
  Our selected parameters are marked with *.}
  \label{tab_ablationCF}
      \begin{tabular}{cc|c|ccc|ccc} \hline
        \multicolumn{2}{c|}{\textbf{Resolution}} &
        \multirow{2}{*}{\textbf{Time(s)}} &
        \multicolumn{3}{c|}{\textbf{Wide}} &
        \multicolumn{3}{c}{\textbf{Narrow}} \\
        \cline{1-2}
        \cline{4-9}
        Coarse & Refinement & & LPIPS$\downarrow$ & SSIM$\uparrow$ & $\boldsymbol{\ell_1(\%)}\downarrow$ & LPIPS$\downarrow$ & SSIM$\uparrow$ & $\boldsymbol{\ell_1(\%)}\downarrow$ \\ 
        \hline
        64* & 256* & 15.3 & \textbf{0.0724} & 0.8834 & \textbf{2.37} & 0.0393 & 0.9174 & 1.20 \\
        128 & 256 & 19.8 & 0.0739 & 0.8832 & 2.48 & 0.0386 & 0.9188 & 1.17 \\
        256 & 256 & 48.6 & 0.0730 & \textbf{0.8837} & 2.38 & \textbf{0.0382} & \textbf{0.9190} & \textbf{1.16} \\
        \hline
      \end{tabular}
  \\
  \vspace{0.1in}
      \begin{tabular}{cc|c|ccc|ccc} \hline
        \multirow{2}{*}{$\boldsymbol{T_c}$} &
        \multirow{2}{*}{$\boldsymbol{T_f}$} &
        \multirow{2}{*}{\textbf{Time(s)}} &
        \multicolumn{3}{c|}{\textbf{Wide}} &
        \multicolumn{3}{c}{\textbf{Narrow}} \\
        \cline{4-9}
        & & & LPIPS$\downarrow$ & SSIM$\uparrow$ & $\boldsymbol{\ell_1(\%)}\downarrow$ & LPIPS$\downarrow$ & SSIM$\uparrow$ & $\boldsymbol{\ell_1(\%)}\downarrow$ \\ 
        \hline
        250* & 75* & 15.3 & 0.0724 & 0.8834 & \textbf{2.37} & \textbf{0.0393} & \textbf{0.9174} & \textbf{1.20} \\
        250 & 50 & 13.4 & 0.0735 & 0.8808 & 2.42 & 0.0410 & 0.9129 & 1.25 \\
        250 & 100 & 18.7 & \textbf{0.0717} & \textbf{0.8820} & 2.44 & 0.0409 & 0.9137 & 1.23 \\
        200 & 75 & 12.6 & 0.0752 & 0.8812 & 2.53 & 0.0405 & 0.9168 & 1.21 \\
        300 & 75 & 20.0 & 0.0747 & 0.8815 & 2.47 & 0.0396 & 0.9164 & 1.22 \\ 
        \hline
      \end{tabular}
\end{table*}

\begin{table*}[t]
  \centering
  \small
  \caption{Effects of Conditioned Resampling Module. 
  }
  \label{tab_ablationRP}
  \begin{tabular}{cc|c|ccc|ccc} \hline
    \multicolumn{2}{c|}{\textbf{CRM}} &
    \multirow{2}{*}{\textbf{Time(s)}} &
    \multicolumn{3}{c|}{\textbf{Wide}} &
    \multicolumn{3}{c}{\textbf{Narrow}} \\
    \cline{1-2}
    \cline{4-9}
    Coarse & Refinement & & LPIPS$\downarrow$ & SSIM$\uparrow$ & $\boldsymbol{\ell_1(\%)}\downarrow$ & LPIPS$\downarrow$ & SSIM$\uparrow$ & $\boldsymbol{\ell_1(\%)}\downarrow$ \\ 
    \hline
    & & 3.0 & 0.1064 & 0.8472 & 3.86 & 0.1016 & 0.8497 & 2.52 \\
    $\checkmark$ & & 8.6 & 0.0758 & 0.8742 & 2.56 & 0.0542 & 0.8935 & 1.49 \\
    & $\checkmark$ & 10.0 & 0.0991 & 0.8591 & 3.65 & 0.0654 & 0.8972 & 1.75 \\
    $\checkmark$ & $\checkmark$ & 15.4 & \textbf{0.0724} & \textbf{0.8834} & \textbf{2.37} & \textbf{0.0393} & \textbf{0.9174} & \textbf{1.20} \\
    \hline
  \end{tabular}
\end{table*}

\renewcommand\arraystretch{1.2}
\begin{table*}[t]
\caption{
Quantitative comparison of our method with the state-of-the-art methods on the test images of CelebA-HQ \cite{karras2017progressive} and ImageNet ~\cite{krizhevsky2012imagenet} with six mask types~\cite{lugmayr2022repaint}. For each metric, the best score is highlighted in \textbf{bold}, and the best score for DDPM-based inpainting methods (\ie  RePaint~\cite{lugmayr2022repaint} and Ours, those are marked with *) is highlighted in \underline{underline}.
}
\label{tab_main}
\centering 
\resizebox{\linewidth}{!}{
  \begin{tabular}{l|c|cccccc|cccccc|cccccc} \hline
\multirow{2}{*}{\textbf{Methods}} & \multirow{2}{*}{\textbf{Dataset}} &
\multicolumn{6}{c|}{\textbf{SSIM}$\uparrow$} & 
\multicolumn{6}{c|}{$\boldsymbol{\ell_1(\%)}\downarrow$} &
\multicolumn{6}{c}{\textbf{LPIPS}$\downarrow$}
\\
\cline{3-20}
& & Wide & Nar. & Half & Exp. & AL & SR
& Wide & Nar. & Half & Exp. & AL & SR
& Wide & Nar. & Half & Exp. & AL & SR
\\\hline
\textbf{DSI \cite{peng2021generating}} & \multirow{8}{*}{\begin{tabular}[c]{@{}c@{}}CelebA\\ -HQ~\cite{karras2017progressive}\end{tabular}} &0.871 & 0.904 & 0.662 & 0.305 & 0.897 & 0.746 & 2.37 & 1.34 & 10.35 & 22.13 & 1.40 & 3.62 & 0.077 & 0.045 & 0.237 & 0.530 & 0.061 & 0.152  \\
\textbf{ICT \cite{wan2021high}} & &0.874 & 0.907 & 0.698 & 0.349 & 0.480 & 0.361 & 2.66 & 1.66 & 9.14 & 22.28 & 10.03 & 8.84 & 0.072 & 0.044 & 0.190 & 0.474 & 0.427 & 0.555  \\
\textbf{MADF \cite{zhu2021image}} & &0.887 & \textbf{0.938} & 0.720 & 0.433 & 0.579 & 0.582 & 2.16 & 1.12 & 7.96 & \textbf{18.30} & 5.90 & 8.77 & 0.076 & 0.042 & 0.214 & 0.475 & 0.389 & 0.347 \\
\textbf{LaMa \cite{suvorov2022resolution}} & &\textbf{0.895} & 0.921 & \textbf{0.749} & \textbf{0.436} & 0.757 & 0.698 & \textbf{1.86} & 1.15 & \textbf{6.73} & 19.24 & 2.45 & 3.75 & \textbf{0.053} & 0.033 & \textbf{0.159} & \textbf{0.414} & 0.111 & 0.225\\
\textbf{MAT \cite{li2022mat}} & &0.886 & 0.921 & 0.712 & 0.359 & 0.672 & 0.238 & 1.91 & \textbf{1.05} & 7.49 & 23.54 & 2.49 & 10.68 & 0.055 & \textbf{0.032} & 0.169 & 0.469 & 0.269 & 0.512 \\
\textbf{RePaint* \cite{lugmayr2022repaint}} & &0.877 & 0.917 & 0.704 & 0.354 & \underline{\textbf{0.975}} & \underline{\textbf{0.926}} & 2.51 & \underline{1.15} & 9.55 & 25.35 & \underline{\textbf{0.50}} & \underline{\textbf{1.13}} & \underline{0.071} & \underline{0.035} & \underline{0.191} & \underline{0.484} & \underline{\textbf{0.010}} & \underline{\textbf{0.032}} \\
\textbf{Ours*} & & \underline{0.887} & \underline{0.918} & \underline{0.724} & \underline{0.402} & 0.952 & 0.900 & \underline{2.32} & 1.20 & \underline{8.66} & \underline{22.23} & 0.74 & 1.37 & 0.072 & 0.039 & \underline{0.191} & 0.490 & 0.026 & 0.053
  \\\hline
\textbf{DSI \cite{peng2021generating}} & \multirow{4}{*}{\begin{tabular}[c]{@{}c@{}}ImageNet\\ ~\cite{krizhevsky2012imagenet}\end{tabular}} & 0.835 & 0.856 & 0.618 & 0.237 & 0.783 & 0.704 & \textbf{3.39} & 2.15 & 10.36 & 23.34 & 2.61 & 4.01 & 0.129 & 0.082 & 0.317 & 0.662 & 0.105 & 0.198  \\
\textbf{ICT \cite{wan2021high}} & &0.833 & 0.848 & 0.645 & \textbf{0.438} & 0.169 & 0.215 & 3.69 & 2.76 & \textbf{9.32} & \textbf{11.37} & 20.66 & 22.58 & \textbf{0.112} & 0.084 & \textbf{0.287} & \textbf{0.422} & 0.615 & 0.651  \\
\textbf{RePaint* \cite{lugmayr2022repaint}} & &0.823 & 0.864 & 0.630 & 0.291 & 0.835 & 0.692 & 4.01 & 2.20 & 11.60 & 26.56 & 2.24 & 3.88 & 0.153 & 0.077 & 0.353 & 0.726 & 0.108 & 0.222   \\
\textbf{Ours*} & & \underline{\textbf{0.842}} & \underline{\textbf{0.870}} & \underline{\textbf{0.657}} & \underline{0.334} & \underline{\textbf{0.842}} & \underline{\textbf{0.736}}
 & \underline{3.46} & \underline{\textbf{1.95}} & \underline{9.55} & \underline{23.74} & \underline{\textbf{2.03}} & \underline{\textbf{3.41}}
 & \underline{0.139} & \underline{\textbf{0.076}} & \underline{0.330} & \underline{0.700} & \underline{\textbf{0.101}} & \underline{\textbf{0.192}} 
  \\\hline
\end{tabular}
}
\end{table*}

\begin{table}[t]
\centering
\caption{
The FID~\cite{heusel2017gans} results on CelebA-HQ \cite{karras2017progressive}. The \textbf{bold} and \underline{underline} represent the best and the second best results.
}
\label{tab_fid}
\small 
\resizebox{\linewidth}{!}{
\begin{tabular}{l|c|c|c|c|c|c} \hline
\multirow{2}{*}{\textbf{Methods}} & \multicolumn{6}{c}{\textbf{FID}$\downarrow$}
\\
\cline{2-7}
& Wide & Nar. & Half & Exp. & AL & SR \\ \hline
\textbf{AOT \cite{zeng2022aggregated}} & 12.19 & 5.63 & 96.98 & 315.52 & 217.40 & 344.52 \\
\textbf{MADF \cite{zhu2021image}} & 9.51 & 7.88 & 26.47 & 142.18 & 86.13 & 82.35 \\
\textbf{LaMa \cite{suvorov2022resolution}} & \textbf{3.46} & 4.28 & 11.44 & 90.48 & \underline{28.42} & 42.14 \\
\textbf{MAT \cite{li2022mat}} & \underline{3.60} & \textbf{3.58} & \textbf{8.87} & \underline{80.06} & 41.13 & 131.95 \\
\textbf{Ours} & 4.14 & \underline{4.08} & \underline{10.73} & \textbf{42.10} & \textbf{7.01} & \textbf{16.60} \\ 
\hline
\end{tabular}
}
\end{table}

\section{Experiments}
\label{sec:exp}
\subsection{Experiment Settings}
\noindent\textbf{Datasets.} We conduct experiments on two datasets (CelebA-HQ~\cite{karras2017progressive} and ImageNet~\cite{krizhevsky2012imagenet}). 
CelebA-HQ dataset contains 30,000 face images at 256$\times$256 pixels. We select the first 27,000 images as train set, and the remaining 3,000 images are set for testing. To show the generalization of our method, we also conduct experiments on the ImageNet dataset.

\vspace{0.05in}
\noindent\textbf{Mask settings.} We use the mask test sets of RePaint~\cite{lugmayr2022repaint} to verify the performance of diverse distributed masks. 
Masks are divided into 6 types: Wide, Narrow, Half, Expand, Alternating Lines and Super-Resolve 2$\times$.

\vspace{0.05in}
\noindent\textbf{Implementation details.} For CelebA-HQ, we employ a lightweight model and conduct training for 500,000 iterations using 27,000 images. 
The hyperparameter settings follow the P2-weighting~\cite{choi2022perception}. 
In testing phase, we evaluate 100 images of size 256$\times$256 from the CelebA-HQ and ImageNet test sets.
We use Structural Similarity (SSIM~\cite{wang2004image}), relative $l1$, and the perceptual metric LPIPS~\cite{zhang2018unreasonable} to evaluate the performance of our method.
The final time expenditure is recorded based on the inference time required for one image on a single RTX 2070 GPU.
Our final configuration is $(T_c, T_f, m_c, m_f, n_c, n_f, s, k)=(250, 75, 3, 2, 8, 10, 5, 2)$, where $m_c, n_c$ are for the coarse stage and $m_f, n_f$ for the refinement stage.

\subsection{Ablation Study}
In this section, we conduct three ablation studies on 100 test images from CelebA-HQ, employing two representative mask types: Wide and Narrow. 
Based on the balanced evaluation of inpainting performance and inference speed, we analyze three speed-up strategies and different designs of CFS. 
In addition, we examine the effects of CRM.

\vspace{0.05in}
\noindent\textbf{Speed-up strategies.}
We conduct experiments to evaluate the effects of speed-up strategies on image inpainting performance. 
The results are shown in Table~\ref{tab_ablation1}. 
We observe LWDM speeds up the process by approximately five times with only a minor impact on the performance. 
While DDIM sampling also provides significant acceleration, it noticeably decreases the inpainting quality. 
CFS not only speeds up the process but also compensates the performance degradation caused by the other two strategies.
For example, comparing between LWDM+DDIM and LWDM+DDIM+CFS, the latter performs better and faster on both Wide and Narrow masks.
Using three speed-up strategies, our method achieves competitive performance with only 1/60 time of baseline.

\vspace{0.05in}
\noindent\textbf{Different designs of Coarse-to-Fine Sampling.}
We redesign CFS by adjusting image resolutions and denoising timesteps $T_c, T_f$. 
Subsequently, we conduct experiments to evaluate their performance across two typical mask types, as illustrated in Table \ref{tab_ablationCF}.
For resolution settings, we choose 64/256 instead of the better-performing 256/256 due to the significantly higher computational cost associated with the latter.
For denoising steps, we set $T_c=250$ and $T_f=75$ to balance the inference speed and the inpainting performance.

\begin{figure}[t]
  \centering
  \includegraphics[width=\linewidth]{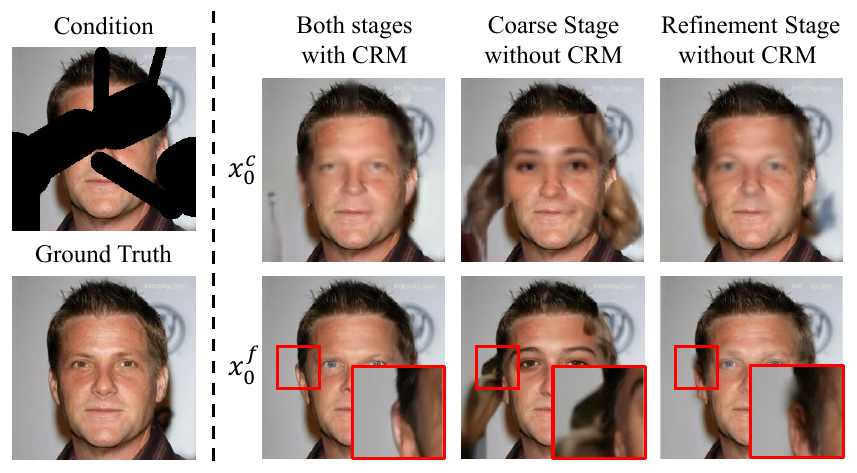}
  \caption{Visualization for the effect of CRM.}  
  \label{fig:RPablation}
\end{figure}

\begin{figure}[t]
  \centering
  \includegraphics[width=1.0\linewidth]{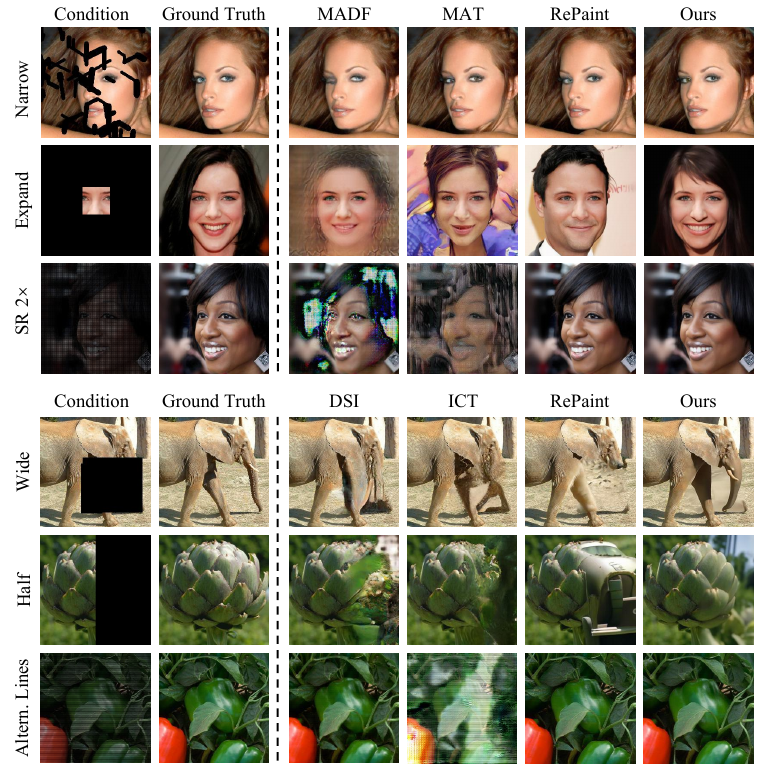}
  \caption{Qualitative results for our method and the state-of-the-arts on the CelebA-HQ~\cite{karras2017progressive} and ImageNet~\cite{krizhevsky2012imagenet} datasets over six different mask types.}
  \label{fig:visualization}
\end{figure}

\vspace{0.05in}
\noindent\textbf{Effects of Conditioned Resampling Module.}
To verify the effectiveness of CRM, we remove it separately from the coarse stage and refinement stage. The results are shown in Table \ref{tab_ablationRP}. We observe that the inpainted result lacks reasonable image structure without coarse stage CRM, and obtains blurred image details without refinement stage CRM, as illustrated in Fig.~\ref{fig:RPablation}.

\subsection{Comparison with the State-of-the-Art Methods}

We compare our method with previous methods on two datasets and list the results in Table~\ref{tab_main}.
Qualitative results are illustrated in Fig. \ref{fig:visualization}.
In CelebA-HQ dataset, our method is competitive with state-of-the-art methods on typical masks~\ie~Wide and Narrow.
Under extreme mask types like Alternating Lines (AL) and Super-Resolve 2$\times$ (SR), our approach notably outperforms other methods not based on DDPM.
Besides, we calculate the Fréchet Inception Distance~\cite{heusel2017gans} (FID) over all 3,000 CeleA-HQ test images and the results are shown in Table~\ref{tab_fid}. 
Our method performs more balanced across different mask types than other methods. 
Espeically for Expand mask inpainting, our method show powerful image generation capabilities.
Without training on specific masks, our method outperforms LaMa~\cite{suvorov2022resolution} across five mask types and achieves competitive performance with MAT.

The additional results are only compared with the methods pre-trained on ImageNet~\cite{krizhevsky2012imagenet}.
With classifier guidance of guided-diffusion~\cite{dhariwal2021diffusion}, our method outperforms RePaint across all masks in any metrics. 

\section{Conclusions}
\label{sec:conclusion}
In this paper, we present an efficient and effective DDPM-based approach for image inpainting, which includes three speed-up strategies. 
Specifically, we replace large-parameter DDPM with Light-Weight Diffusion Model (LWDM) combined with training objective modified.
And we substitute DDPM sampling with skip-step DDIM sampling to accelerate the denoising process.
Furthermore, we propose Coarse-to-Fine Sampling (CFS) strategy to further speed up and improve the performance.
Experimental results on facial and general-purpose image inpainting tasks demonstrate that our method achieves competitive results across different mask types with approximately 60$\times$ speedup compared with RePaint~\cite{lugmayr2022repaint}.



\small{
  \bibliographystyle{IEEEbib}
  \bibliography{total}
}

\end{document}